\documentclass[11pt]{article} % For LaTeX2e
\usepackage{rldmsubmit,palatino}
\usepackage{amsmath}
\usepackage{graphics,graphicx,color}	%figures and color
\usepackage[pdftex]{epsfig}
\usepackage{caption}
\usepackage{subcaption}
\usepackage{wrapfig}
\usepackage{makecell}
\usepackage{color}
\title{Policy Learning with Hypothesis based Local Action Selection}
\author{
Bharath ~Sankaran\thanks{ http://www-clmc.usc.edu/Main/BharathSankaran} \\
Department of Computer Science\\
University of Southern California\\
Los Angeles, CA 90089 \\
\texttt{bsankara@usc.edu} \\
\And
Jeannette Bohg \\
Max Planck Institute for Intelligent Systems \\
T{\"u}bingen, Germany 72076  \\
\texttt{jeannette.bohg@tuebingen.mpg.de } \\
\AND
Nathan Ratliff \\
Max Planck Institute for Intelligent Systems \\
University of Stuttgart\\
\texttt{nathan.ratliff@tuebingen.mpg.de} \\
\And
Stefan Schaal \\
University of Southern California \\
Max Planck Institute for Intelligent Systems\\
\texttt{sschaal@usc.edu} \\
}

\begin{document}

\maketitle

\begin{abstract}

For robots to be effective in human environments, they should be capable of successful task execution in unstructured environments. Of these, many task oriented manipulation behaviors executed by robots rely on model based grasping strategies and  model based strategies require accurate object detection and pose estimation. Both these tasks are hard in human environment, since human environments are plagued by partial observability and unknown objects.  Given these constraints, it becomes crucial for a robot to be able to operate effectively under partial observability in unrecognized environments. Manipulation in such environments is also particularly hard, since the robot needs to reason about the dynamics of how various objects of unknown or only partially known shape interact with each other under contact. Modelling the dynamic process of a cluttered scene during manipulation is hard even if all object models and poses were known. It becomes even harder to reasonably develop a process or observation model, with only partial information about the object class or shape. To enable a robot to effectively operate in \emph{partially observable unknown environments} we introduce a policy learning framework where action selection is cast as a \emph{probabilistic classification problem} on hypothesis sets generated from observations of the environment. Online the action classifier is operated with a global stopping criterion for successful task completion. The example we consider is object search in clutter, where we assume having access to a visual object detector, that directly populates the hypothesis set given the current observation. Thereby we can avoid the temporal modelling of the process of searching through clutter. We demonstrate our algorithm on two manipulation based object search scenarios; a \emph{modified minesweeper simulation} and a \emph{real world object search in clutter} using a dual arm manipulation platform.
\end{abstract}


\acknowledgements{This research was supported in part by National Science Foundation grants IIS-1205249, IIS-1017134, CNS-0960061, EECS-0926052, the DARPA program on Autonomous Robotic Manipulation, the Office of Naval Research, the Okawa Foundation, and the Max-Planck-Society.}

\startmain % to start the main 1-4 pages of the submission.
\section{Introduction}
\label{sec:Introduction}
For robots to be able to manipulate in unknown and unstructured environments the robot should be capable of operating under partial observability of the environment. Object occlusions and unmodeled environments are some of the factors that result in partial observability which in turn causes an uncertainty in the robot state estimate.
A common scenario where this is encountered is \emph{manipulation in clutter}. In the case that the robot needs to locate an object of interest and manipulate it, it needs to perform a series of decluttering actions to accurately detect the object of interest.
To perform such a series of actions, the robot also needs to account for the dynamics of objects in the environment and how they react to contact. This is a non trivial problem since one needs to reason not only about robot-object interactions but also object-object interactions in the presence of contact. 
In the example scenario of \emph{manipulation in clutter}, the state vector would have to account for the pose of the object of interest and the structure of the surrounding environment. The process model would have to account for all the aforementioned robot-object, object-object interactions. 
The complexity of the process model grows exponentially as the number of objects in the scene increases. This is commonly the case in unstructured environments. Hence it is not reasonable to attempt to model all object-object and robot-object interactions explicitly.

Also in some cases of human decision making we observe that we don't reason over all the possible agent-object and object-object interactions when manipulating in unstructured environments. For instance, imagine the case where you are looking for your keys on a table among clutter. When sifting through clutter we don't reason about all possible agent-object or object-object interactions. Since we have an accurate model of the object of interest, i.e the keys, we only reason about a limited set of cases. Such as the 
 possibility of the keys being occluded by an object, etc. Under this setting we can formulate the problem as one where we construct a set of hypothesis about the possible poses of the object of interest given the current evidence in the scene and select actions based on our current set of hypothesis. This hypothesis set tends to represent the belief about the structure of the environment and the number of poses the object of interest can take. The uncertainty relating to the pose of the object of interest is directly dependent on the structure of the environment, i.e on the number other known or unknown objects in the environment. 
The agent's only stopping criterion is when the uncertainty regarding the pose of the object is \emph{fully resolved}. The question to naturally pose is, is it possible to learn a search policy for such settings in real systems. Also what are the constraints that must be applied to the problem setting to make learning tractable. A crucial factor to note is, as the size of the environment grows, the size of this hypothesis set also grows. 
%In the following sections we formulate our policy learning problem for partially observable environments and demonstrate why extensively modeling the problem is not tractable. We then introduce our solution to action selection, where we recast the policy learning problem as a hypothesis set based action classification problem. 
%We then demonstrate our policy learning framework on hidden object search problems with experiments in a modified minesweeper simulation environment and how this can be extended to a real robot decluttering experiment. We also demonstrate how our approach generalizes trivially to growing state spaces.
%If formulated as a Partially Observable Markov Decision Process (POMDP), It is non-trivial to learn a transition/process model for these hypothesis updates since the size of the set of hypothesis might change depending on what robot-object, object-object interactions occur.

\section{Problem Formulation}
\label{sec:problem_formulation}
Consider a robot that has access to a database of object models $\mathcal{O} = \{O_1,....,O_n\}$ and a set of actions $\mathcal{A} = \{a_1,...,a_K\}$. These actions could be movement primitives. Our task is to locate an object of interest $O_i\in\mathcal{O}$ in a cluttered environment. To accomplish this task, we need to execute a sequence of actions from $\mathcal{A}$ to manipulate the environment, to accurately detect $\mathcal{O}_i$. 
For this problem we denote our current state vector as $X_t \in \mathcal{X}$ which comprises of the pose of $\mathcal{O}_i$ represented by $\mathcal{P}_t \in \mathcal{P}$. $\mathcal{P}_t$ is dictated by an object model and the current structure of the environment $\mathcal{E}_t \in \mathcal{E}$. $\mathcal{E}_t$ is a voxelized representation where the occupancy of voxels are informed by the poses of all the other detected objects in the environment, whose shapes are dictated by object models or shape primitives.
Let $b$ denote the belief state, i.e. the distribution over the state space $\mathcal{X}$. Our objective is to learn a policy that will give us an action to execute given our current belief about the state. In essence we want to learn a policy 
$\pi: b(X_t) \rightarrow \mathcal{A}$, where $X_t = [\mathcal{P}_t; \mathcal{E}_t]$. To determine the optimal sequence of actions to achieve our task, we can formulate the problem as a POMDP, where our optimal policy would be given by
\[
 \pi^* = \underset{\pi} {\mathrm{argmax}} \text{ }  V^\pi(b(X_0))
\]
where $b(\mathcal{X}_0)$ is our initial belief.
The optimal policy, denoted by $\pi^*$ yields the highest expected reward value for each belief state, which is represented by an optimal value function $V^*$. This value function can be calculated as 
\[
V^*(b(X_t)) = \underset{a\in\mathcal{A}} {\mathrm{max}} \text{ }\left[ \mathcal{R}(b(X_t), a) + \gamma{\mathrm{\sum\limits_{Z_{t}\in\mathcal{Z}}}} O(Z_{t}| b(X_{t}), a)V^*(\tau(b(X_{t}),a,Z_{t})) \right]  
\]

Here $\gamma$ is a discount factor and our reward is defined as:
\[\mathcal{R}(b(X_t), a) = \left\{
  \begin{array}{lr}
    1 & \text{if } a=a_{\text{ter}}\\
    0 & \text{otherwise}
  \end{array}
\right.
\]
An action $a = a_{\text{ter}}$ if the object of interest is successfully located.
In this formulation we also assume access to an observation model $O(Z_{t}| b(X_{t}), a)$ and a process model $\tau(b(X_{t}),a,Z_{t})$, i.e we can accurately predict the outcome of an action. The process model in this formulation inherently assumes one of two criteria. Either we can model the dynamics of interactions between various rigid bodies in the environment or we can model the evolution of the hypothesis set as an outcome of actions executed. 
As mentioned earlier in Section \ref{sec:Introduction}, both of these tasks are non trivial. Given the context of our problem it is not easy to model object-object and robot-object interactions \emph{or} model the change in the state uncertainty as an outcome of physical interaction. A possible argument to model either of these phenomena would be to learn from demonstrations or synthetic data.
Even if we were to learn these distributions from demonstrations or synthetic data, the number of samples required to reasonably approximate the state space would be exponential in the number of objects in the environment. A similar argument can be made for the observation model.
%\[
%\vec{\mathcal{X}}_t = \left\{
%  \begin{array}{c}
%    \mathcal{P}_t\\
%    \mathcal{E}_t
%  \end{array}
%\right\}
%\]
Also, the belief function $b(X_t)$ is hard to estimate given a large state space, as it needs to account for the object pose $\mathcal{P}$ and the entire structure of the environment $\mathcal{E}$. Hence, we constrain this general formulation. 

We note that we can in principle filter the belief using Bayesian filtering to account for the entire history of observations and actions. In our case, the belief function $b()$ represents the distribution over the object poses and structure of the environment. Note that the object poses are dependent on the structure of the environment hence modeling this uncertainty is not straightforward. Instead of parameterizing the distribution of the state vector $X_t$, we adopt a non parameterized approach where we use a discrete set of hypotheses $\mathcal{H} = \{H_1,....,H_m\}$ that can be constructed using the model of our object of interest $\mathcal{O}_i$ and the current state of the environment $\mathcal{E}_t$. 
The state of the environment at time $t$ is estimated from observation $Z_t$ given by a visual sensor. Given our current observation $Z_t$, we specify the belief $b(X_t)$ as the current hypotheses object poses with respect to the visible environment, given by the set $\mathcal{H}_t = b(X_t)$. This hypothesis set is constructed using tools from vision that take the object model $\mathcal{O}_i$ and observation $Z_t$ and return $\mathcal{H}_t = \phi(Z_t,\mathcal{O}_i)$. 
The objective of the problem is to manipulate the environment till we have reduced the cardinality of our current hypothesis set to 1, $\|\mathcal{H}_t\| = 1$ so that we can successfully execute a model based manipulation action. We define this action as a terminal action $a_{ter}\in \mathcal{A}$ with reward 1. 
In an effort to make learning and inference in this setting tractable, we approximate quantities that can easily observed and modeled. Instead of trying to learn the dynamics of interactions in the environment, we try to directly learn a mapping between the belief state $b(X_t)$ and actions $\mathcal{A}$. This mapping is learned with discriminative classifiers that return an action given the current belief state.
To ensure that the state space of the problem does not grow exponentially with the number of objects in the scene, we make the classifiers agnostic to the complete state of the environment and instead have them classify actions based on features computed on the current hypothesis set $\mathcal{H}_t$. We assume that we can construct the hypothesis set for any object model $\mathcal{O}$ under any observation in $\mathcal{Z}$, i.e $\mathcal{H} = \phi(Z_t,\mathcal{O}_i)$.
Hence our policy learning problem is reduced to
\[
  \pi^* = \underset{a} {\mathrm{argmax}} \text{ }  w^Tf(b(X_t),a) \text{ where }b(X_t) = \mathcal{H}_t 
\]
Here different policies can be learned and compared by either altering the features or the number of classes, i.e actions.
\section{Modified Minesweeper Simulation}
We emulate the problem of action selection under partial observability using a modified minesweeper scenario. In our modified minesweeper scenario, the mines are organized into a fixed size H-structure in the grid. The objective of the game is to accurately determine the pose of this hidden H-structure by opening a minimum number of non-mine cells. 
\begin{wrapfigure}{r}{0.35\textwidth}
  \vspace{-15pt}
  \begin{center}
    \begin{subfigure}[b]{0.10\textwidth}
        \centering
        \includegraphics[width=\textwidth,height=20mm]{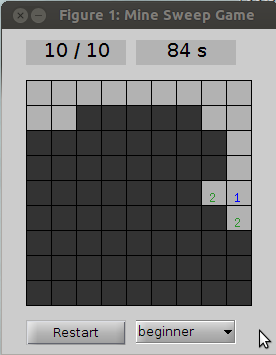}
        \caption{}
        %\caption{Game play\\ environment}
        \label{fig:game_play}
    \end{subfigure}
    \hfill
    \begin{subfigure}[b]{0.10\textwidth}
        \centering
        \includegraphics[width=\textwidth,height=20mm]{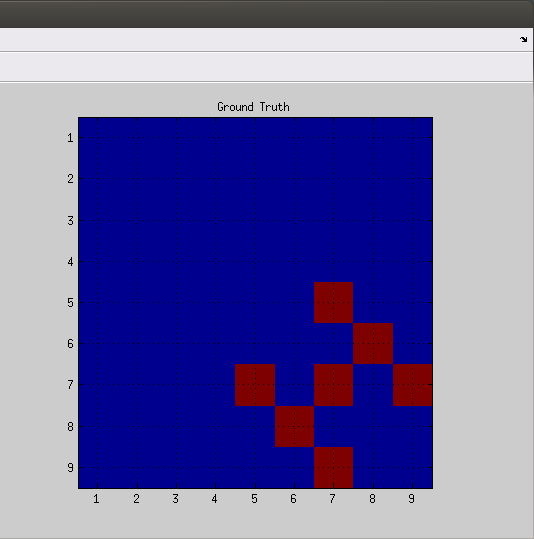}
        \caption{}
        %\caption{Hidden H-structure}
        \label{fig:hidden_I}
    \end{subfigure}    
    \hfill
    \begin{subfigure}[b]{0.10\textwidth}
        \centering
        \includegraphics[width=\textwidth,height=20mm]{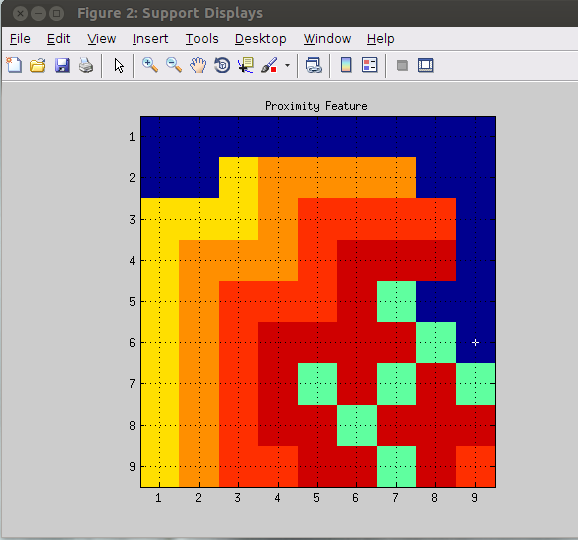}
        \caption{}
        %\caption{Hypothesis Set with cardinality 1}
        \label{fig:hypothesis_set}
    \end{subfigure}    
        \caption{Modified Minesweeper}
\label{fig:modified_minesweeper}
  \end{center}
  \vspace{-10pt}
\end{wrapfigure}
As in the classical minesweeper scenario opened cells may either be numbered or empty indicating the number of mines in the 8-connected neighbourhood \emph{or} the opened cell might be a mine in which case the game terminates.
The agent selects actions based on its current hypothesis set. This set is constructed based on the current observation, i.e opened cells and their values. The game is completed when the agent has narrowed down its set of hypothesis to one. The set of actions available to the agent is to open a cell from the 8-connected neighbourhood of the current open cell. The game play is initialized randomly.
\begin{wrapfigure}{r}{0.35\textwidth}
  \vspace{-15pt}
  \begin{center}
        \begin{subfigure}[b]{0.10\textwidth}
        \centering
        \includegraphics[width=\textwidth,height=20mm]{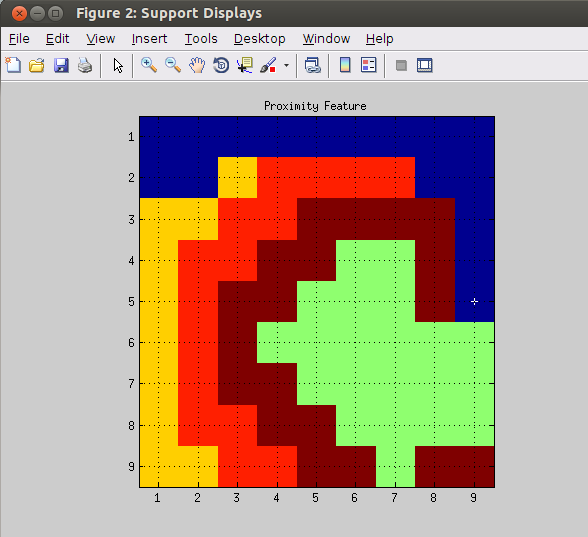}
        \caption{}
        \label{fig:hyp_evolution4}
    \end{subfigure}
    \hfill
    \begin{subfigure}[b]{0.10\textwidth}
        \centering
        \includegraphics[width=\textwidth,height=20mm]{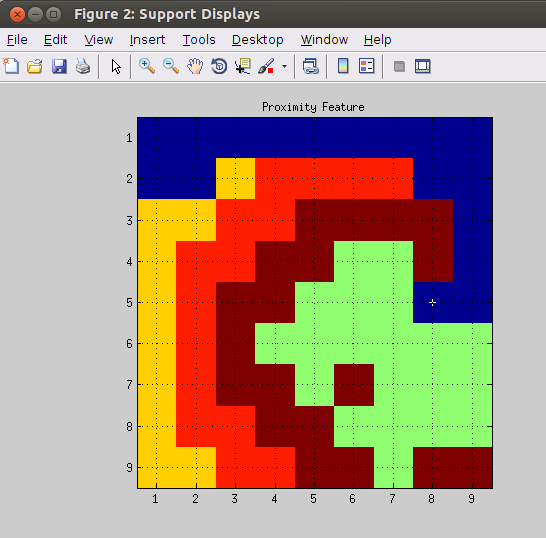}
        \caption{}
        \label{fig:hyp_evolution5}
    \end{subfigure}    
    \hfill
    \begin{subfigure}[b]{0.10\textwidth}
        \centering
        \includegraphics[width=\textwidth,height=20mm]{minesweeper_hypothesis3.png}        
        \caption{}
        \label{fig:hyp_evolution6}
    \end{subfigure}      
  \end{center}
  \vspace{-10pt}
%\end{wrapfigure}
%\begin{wrapfigure}{r}{0.35\textwidth}
  \vspace{-10pt}
  \begin{center}
        \begin{subfigure}[b]{0.10\textwidth}
        \centering
        \includegraphics[width=\textwidth,height=20mm]{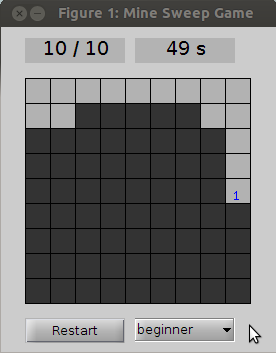}
        \caption{}
        \label{fig:hyp_game4}
    \end{subfigure}
    \hfill
    \begin{subfigure}[b]{0.10\textwidth}
        \centering
        \includegraphics[width=\textwidth,height=20mm]{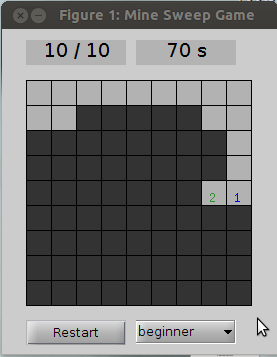}
        \caption{}
        \label{fig:hyp_game5}
    \end{subfigure}    
    \hfill
    \begin{subfigure}[b]{0.10\textwidth}
        \centering
        \includegraphics[width=\textwidth,height=20mm]{minesweeper_game3.png}        
        \caption{}
        \label{fig:hyp_game6}
    \end{subfigure}  
\caption{Hypothesis Set and Game Environment Updates}
    \label{fig:hyp_evolution}
  \end{center}
  \vspace{-20pt}
\end{wrapfigure}

A demonstration of this game play environment is show in Figure \ref{fig:modified_minesweeper}, where Figure \ref{fig:game_play} is the actual game play environment, Figure \ref{fig:hidden_I} is the ground truth location of the hidden H-structure and Figure \ref{fig:hypothesis_set} shows the features computed on the current hypothesis set.
The feature we use is an inverse distance transform where cells close to the current set of hypothesis get a high score and cells far away from the hypothesis set get a low score. We then extract local templates from the features computed on the hypothesis set. These templates are 3x3 patches around the current expert location. The class corresponding to the feature is the location of the next action selected by the expert in the 8-connected neighbourhood.  The evolution of the hypothesis set corresponding to the current game environment is demonstrated in Figure \ref{fig:hyp_evolution}.
We train the agent with demonstrations from an expert where the expert plays the game over a number of trials. 
\begin{wraptable}{r}{0.55\textwidth}
\tiny
  \begin{center} 
    \begin{tabular}{| l | l | l | l | l | l | l | l | l | l | l |}
    \hline
    Agent&Trial 1&Trial 2&Trial 3&Trial 4&Trial 5&Trial 6&Trial 7&Trial 8&Trial 9&Trial  10\\    \hline
    \textbf{MC} & \colorbox{green}{12.3} & \colorbox{green}{8.4;\textbf{2}} & \colorbox{green}{8.1} & 10.1 & \colorbox{green}{7.1;\textbf{1}} & \colorbox{green}{8.3} & \colorbox{green}{7.1;\textbf{1}} & 10.1 & 12.6 & \colorbox{green}{13.2;\textbf{2}}\\
    \hline
    \textbf{BE} & 13.8 & 11.3 & 13.2 & 16 & 10.6 & 15.6 & 20.8 & 11.6 & \colorbox{green}{12.1} & 14.5  \\
    \hline
    \textbf{B8} & 12.3;\textbf{3}& \colorbox{green}{8.4;\textbf{2}} & 9.1;\textbf{3}  & 10.1;\textbf{2} & 9;\textbf{4} & 17.8;\textbf{4} & 8.3;\textbf{3} & 10.1;\textbf{2} & 12.6;\textbf{5} & 13.2;\textbf{3}\\
    \hline
    \textbf{HP} & 25.6 & 9.3 & 8;\textbf{1} & \colorbox{green}{6.4} & 7.8;\textbf{1} & 13.8 & 8.3 & \colorbox{green}{7.1} & 28.5 & 21\\
    \hline
    \end{tabular}    
  \end{center}  
  \caption{Results of Minesweeper Tests} \label{tab:mine_results}
\end{wraptable} 

We compare different agents against a heuristic player (\textbf{HP}). The agents trained were a Multiclass (\textbf{MC}) 1 vs all SVM trained on the local templates with 8-connected neighbourhood as; a binary agent (\textbf{BE}) that classifies a local template from anywhere on the grid as actionable or not and a binary 8-connected (\textbf{B8}) agent that applies the binary agent to the 8-connected grid.
We tested the various agents over 100 different trials with 10 random poses of the hidden H-structure and each of the 10 poses had 10 different initializations for the agent. The results are tabulated in Table \ref{tab:mine_results}. The results show the mean number of actions taken over the successful trials off the 10 trials. The number of failed attempts in these 10 trials are boldfaced. Failures result due to opening a mine or in the \textbf{B8} case failing to classify any neigbouring grid as actionable. The best result for each random pose are highlighted in green.

\section{Transition to a Real Robot Environment}
We apply the same policy learning framework to a real robot decluttering experiment, where the robot is tasked with locating an object of interest in a cluttered environment. Here the input observation $\mathcal{Z}_t$ is an RGBD pointcloud. The hypothesis set $\mathcal{H}_t$, of the object of interest is computed using the output of an object classifier \cite{SankaranTR014}, that returns an object class and pose hypothesis for every pointcloud cluster in the environment. These hypotheses are then projected on to a planar support surface (tabletop) to compute a hypothesis feature similar to the minesweeper scenario.
The general pipeline is demonstrated in the figure below.
\begin{figure}[hb!]
    \centering
    \begin{subfigure}[b]{0.20\textwidth}
        \centering
        \includegraphics[width=\textwidth,height=18mm]{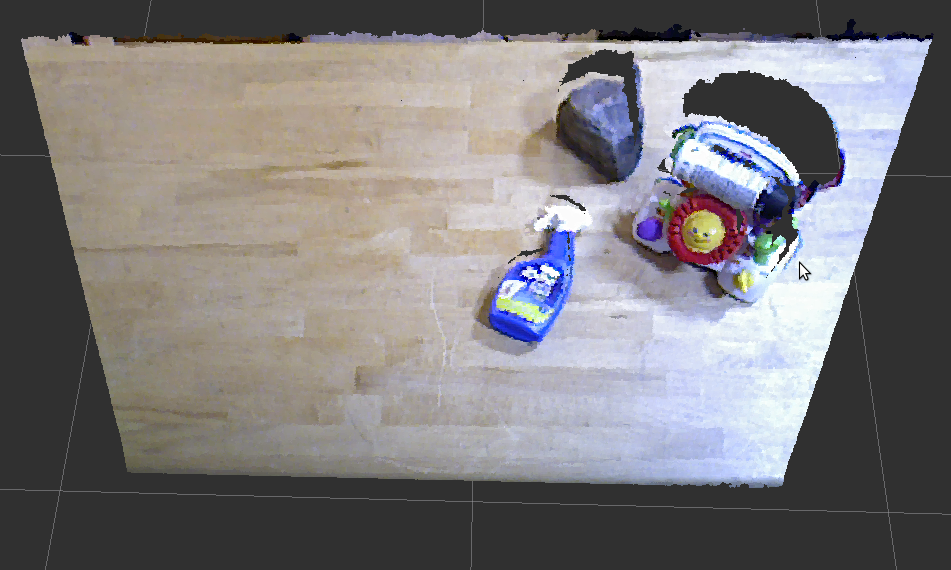}
        \caption{Input Point Cloud}
        \label{fig:input_pointlcoud}
    \end{subfigure}
    \hfill
    \begin{subfigure}[b]{0.20\textwidth}
        \centering
        \includegraphics[width=\textwidth,height=18mm]{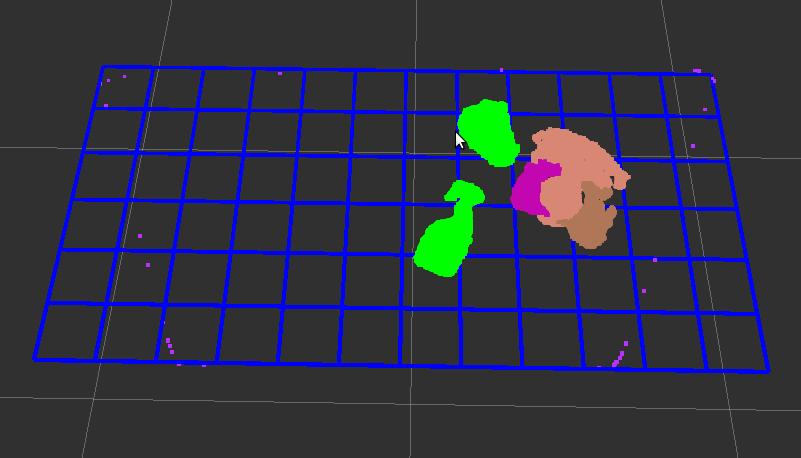}
        \caption{Pointcloud Clustering}
        \label{fig:clustering}
    \end{subfigure}    
    \hfill
    \begin{subfigure}[b]{0.20\textwidth}
        \centering
        \includegraphics[width=\textwidth,height=18mm]{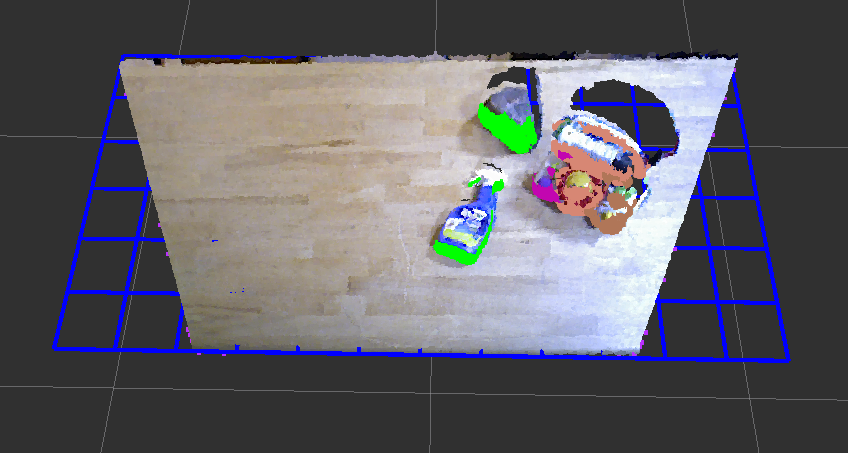}
        \caption{Preprocessing Overlay}
        \label{fig:preprocessing_overlay}
    \end{subfigure}  
    \hfill
        \begin{subfigure}[b]{0.20\textwidth}
        \centering
        \includegraphics[width=\textwidth,height=18mm]{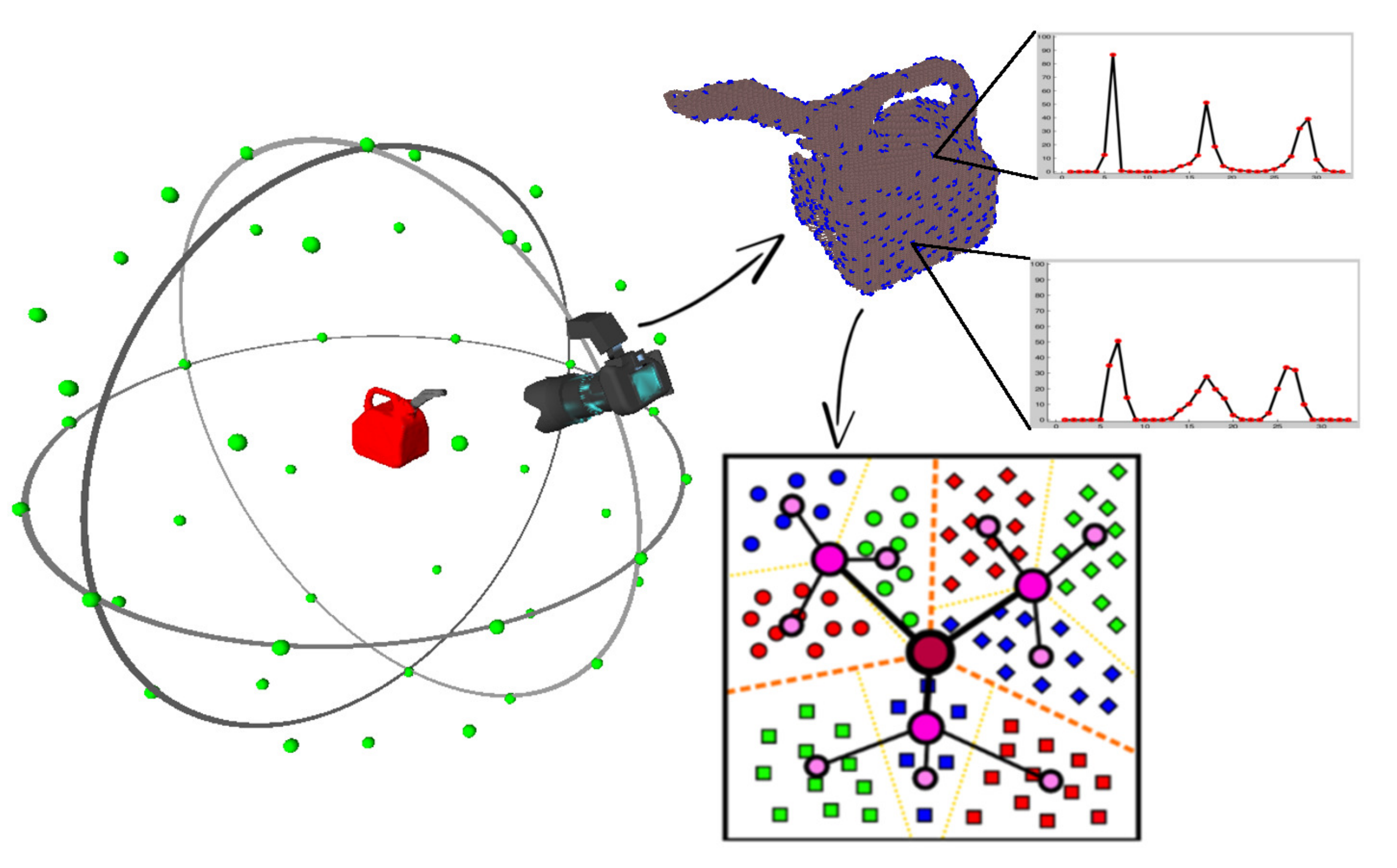}
        \caption{VP-Tree Classifier}
        \label{fig:vp_tree}
    \end{subfigure}
    \caption{Point cloud preprocessing}
    \label{fig:pointcloud_preprocessing}
\end{figure}
\vspace{-15pt}
\begin{figure}[hb!]
    \centering
    \begin{subfigure}[b]{0.20\textwidth}
        \centering
        \includegraphics[width=\textwidth,height=18mm]{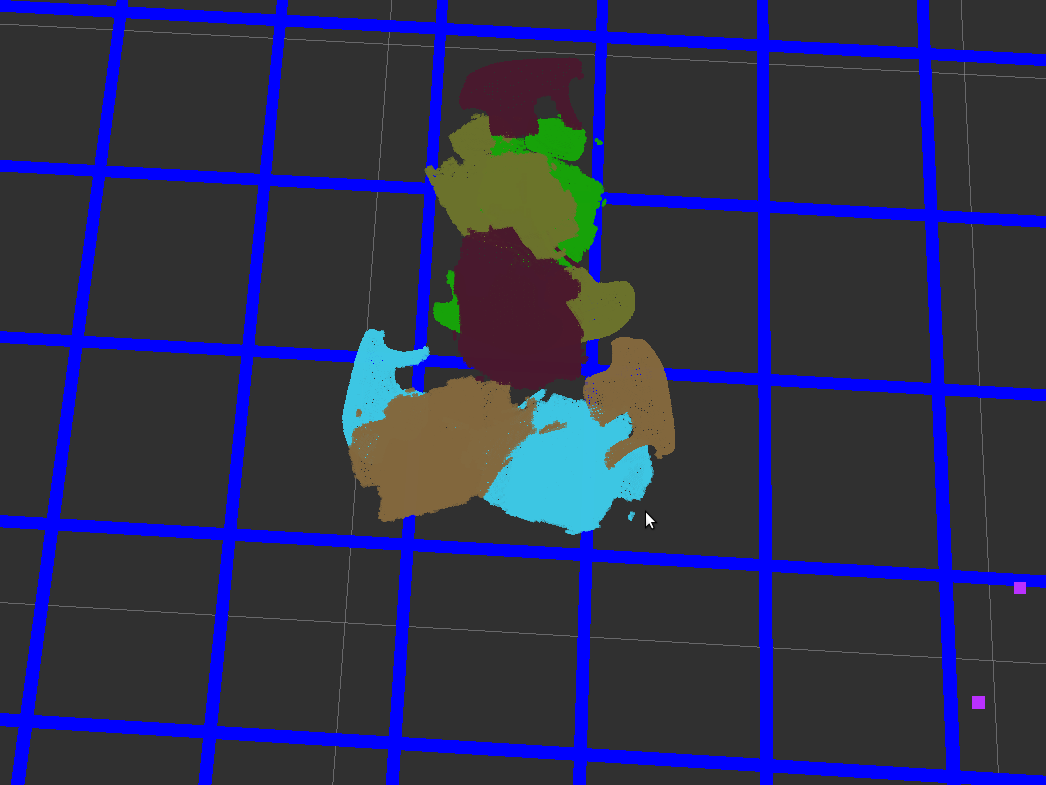}
        \caption{Projected hypothesis}
        \label{fig:proj_hypothesis}
    \end{subfigure}
    \hfill
    \begin{subfigure}[b]{0.20\textwidth}
        \centering
        \includegraphics[width=\textwidth,height=18mm]{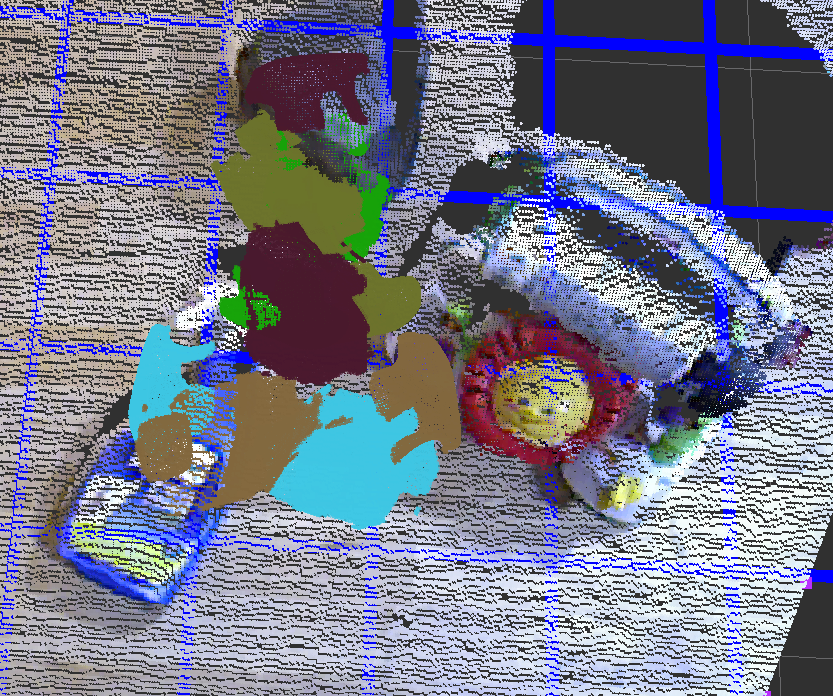}
        \caption{Hypothesis Overlay}
        \label{fig:hyp_overlay}
    \end{subfigure}    
    \hfill
    \begin{subfigure}[b]{0.20\textwidth}
        \centering
        \includegraphics[width=\textwidth,height=18mm]{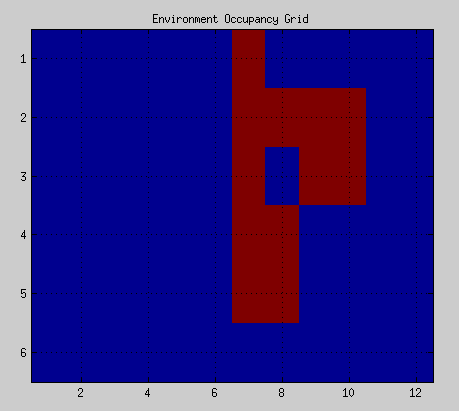}
        \caption{Env Occupancy Grid}
        \label{fig:occupancy_grid}
    \end{subfigure}  
        \hfill
    \begin{subfigure}[b]{0.20\textwidth}
        \centering
        \includegraphics[width=\textwidth,height=18mm]{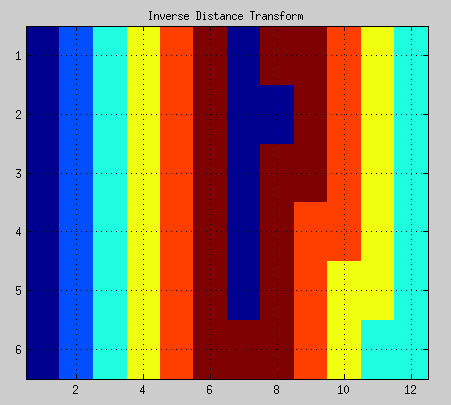}
        \caption{Inverse Dist Transform}
        \label{fig:inverse_dist_robot}
    \end{subfigure}  
    \caption{Hypothesis Feature Computation}
    \label{fig:hypothesis_feature}
    \vspace{-15pt}
\end{figure}
\section{Conclusions and Future Work}
We have demonstrated a policy learning approach for hypotheses based action selection. Our approach is trained in a supervised manner with expert demonstrations. The key features of our approach are we can accomplish complex tasks without reasoning about a process or observation model. Our approach also has the ability to scale to large environments and the learning complexity is agnostic to the size of the environment. Our proposed model simplification approach is only valid for the class of POMDP problems where states are strictly markovian in nature ex: \cite{Ross_2011,Ross_2010}, i.e where the current observation encompases the history of all previous observations. In the future we are going to perform more tests on our robotic setup and apply this frame work to other policy learning tasks.
% 
% 
% Authors preferring \LaTeX{} are requested to use the RLDM \LaTeX{}
% style files obtainable at the RLDM website at
% \begin{center}
%    http://www.rldm.org/
% \end{center}
% The file \verb+rldm.pdf+ contains these instructions and illustrates
% the various formatting requirements your RLDM paper must
% satisfy. There is a \LaTeX{} style file called \verb+rldmsubmit.sty+,
% and a \LaTeX{} file \verb+rldm.tex+, which may be used as a ``shell''
% for writing your paper. All you have to do is replace the author,
% title, abstract, keywords, acknowledgements and text of the paper with
% your own. The file
% \verb+rldm.rtf+ is provided as an equivalent shell for Microsoft Word users. 

% \section{General formatting instructions}
% \label{gen_inst}
% 
% The paper size for RLDM is ``US Letter'' (rather than ``A4''). Margins
% are 1.5cm around all sides. Use 11~point type with a vertical spacing
% of 12~points. Palatino is the preferred typeface throughout.
% Paragraphs are separated by 1/2~line space, with no indentation.
% 
% Paper title is 17~point, initial caps/lower case, bold, centered between
% 2~horizontal rules. Top rule is 4~points thick and bottom rule is 1~point
% thick. Allow 0.6cm space above and below title to rules. 
% 
% The lead author's name is to be listed first (left-most), and
% the co-authors' names (if different address) are set to follow. If
% there is only one co-author, list both author and co-author side by side.
% 
% \section{Preparing PostScript or PDF files}
% 
% Please prepare PostScript or PDF files with paper size ``US Letter''.
% The -t letter option on dvips will produce US Letter files.
\nocite{*}
%==================================================================%
\tiny
\bibliographystyle{IEEEtran}
\bibliography{ipreview_Bibliography}

% Generated by IEEEtran.bst, version: 1.13 (2008/09/30)
\begin{thebibliography}{1}
\providecommand{\url}[1]{#1}
\csname url@samestyle\endcsname
\providecommand{\newblock}{\relax}
\providecommand{\bibinfo}[2]{#2}
\providecommand{\BIBentrySTDinterwordspacing}{\spaceskip=0pt\relax}
\providecommand{\BIBentryALTinterwordstretchfactor}{4}
\providecommand{\BIBentryALTinterwordspacing}{\spaceskip=\fontdimen2\font plus
\BIBentryALTinterwordstretchfactor\fontdimen3\font minus
  \fontdimen4\font\relax}
\providecommand{\BIBforeignlanguage}[2]{{%
\expandafter\ifx\csname l@#1\endcsname\relax
\typeout{** WARNING: IEEEtran.bst: No hyphenation pattern has been}%
\typeout{** loaded for the language `#1'. Using the pattern for}%
\typeout{** the default language instead.}%
\else
\language=\csname l@#1\endcsname
\fi
#2}}
\providecommand{\BIBdecl}{\relax}
\BIBdecl

\bibitem{SankaranTR014}
N.~Atanasov, B.~Sankaran, J.~Le~Ny, G.~Pappas, and K.~Daniilidis, ``Nonmyopic
  view planning for active object classification and pose estimation,'' 2014.

\bibitem{Ross_2011}
S.~Ross, G.~Gordon, and J.~A.~D. Bagnell, ``A reduction of imitation learning
  and structured prediction to no-regret online learning,'' in
  \emph{Proceedings of the 14th International Conference on Artifical
  Intelligence and Statistics (AISTATS)}, April 2011.

\bibitem{Ross_2010}
S.~Ross and J.~A.~D. Bagnell, ``Efficient reductions for imitation learning,''
  in \emph{Proceedings of the 13th International Conference on Artificial
  Intelligence and Statistics (AISTATS)}, May 2010.

\bibitem{NgIRLICML00}
\BIBentryALTinterwordspacing
A.~Y. Ng and S.~J. Russell, ``Algorithms for inverse reinforcement learning,''
  in \emph{Proceedings of the Seventeenth International Conference on Machine
  Learning}, ser. ICML '00.\hskip 1em plus 0.5em minus 0.4em\relax San
  Francisco, CA, USA: Morgan Kaufmann Publishers Inc., 2000, pp. 663--670.
  [Online]. Available: \url{http://dl.acm.org/citation.cfm?id=645529.657801}
\BIBentrySTDinterwordspacing

\bibitem{DogarICRA13}
M.~R. Dogar, M.~C. Koval, A.~Tallavajhula, and S.~S. Srinivasa, ``Object search
  by manipulation,'' in \emph{ICRA}, 2013, pp. 4973--4980.

\bibitem{Crammer:2002}
\BIBentryALTinterwordspacing
K.~Crammer and Y.~Singer, ``On the algorithmic implementation of multiclass
  kernel-based vector machines,'' \emph{J. Mach. Learn. Res.}, vol.~2, pp.
  265--292, Mar. 2002. [Online]. Available:
  \url{http://dl.acm.org/citation.cfm?id=944790.944813}
\BIBentrySTDinterwordspacing

\end{thebibliography}
%==================================================================%

\end{document}